\documentclass[conference]{IEEEtran}
\IEEEoverridecommandlockouts

\usepackage{cite}
\usepackage{amsmath,amssymb,amsfonts}
\usepackage{algorithmic}
\usepackage{graphicx}
\usepackage{textcomp}
\usepackage{xcolor}
\usepackage{multicol}

\usepackage{tikz}
\newcommand\copyrighttext{
    \footnotesize \textcopyright~2023 IEEE. Personal use of this material is permitted. Permission from IEEE must be obtained for all other uses, in any current or future media, including reprinting/republishing this material for advertising or promotional purposes, creating new collective works, for resale or redistribution to servers or lists, or reuse of any copyrighted component of this work in other works. The final version of this article is available at: https://doi.org/10.1109/ICAICTA59291.2023.10390269
}
\newcommand\copyrightnotice{
    \begin{tikzpicture}[remember picture, overlay]
        \node[anchor=south, yshift=10pt] at (current page.south) {
            \fbox{
                \parbox{\dimexpr1.0\textwidth-\fboxsep-\fboxrule\relax}{\copyrighttext}
            }
        };
    \end{tikzpicture}
}

\PassOptionsToPackage{hyphens}{url}\usepackage{hyperref}

\def\BibTeX{{\rm B\kern-.05em{\sc i\kern-.025em b}\kern-.08em
    T\kern-.1667em\lower.7ex\hbox{E}\kern-.125emX}}
\begin{document}

\title{
    Simple Hack for Transformers \\ against Heavy Long-Text Classification \\ on a Time- and Memory-Limited GPU Service
}

\author{
    \IEEEauthorblockN{
        Mirza Alim Mutasodirin\IEEEauthorrefmark{1}, 
        Radityo Eko Prasojo\IEEEauthorrefmark{2}, 
        Achmad F. Abka\IEEEauthorrefmark{2}\IEEEauthorrefmark{3},
        Hanif Rasyidi\IEEEauthorrefmark{4}
    }
    
    \IEEEauthorblockA{\IEEEauthorrefmark{1}Politeknik Harapan Bersama, Tegal, Indonesia}
    \IEEEauthorblockA{\IEEEauthorrefmark{2}Universitas Indonesia, Depok, Indonesia}
    
    \IEEEauthorblockA{\IEEEauthorrefmark{3}National Research and Innovation Agency, Cibinong, Indonesia}
    \IEEEauthorblockA{\IEEEauthorrefmark{4}Australian National University, Canberra, Australia}
    
    \IEEEauthorblockA{
        \IEEEauthorrefmark{1}mirza.alim.m@poltektegal.ac.id,
        \IEEEauthorrefmark{2}radityo.ep@ui.ac.id,
        \IEEEauthorrefmark{3}achm034@brin.go.id
        \IEEEauthorrefmark{4}hanif.rasyidi@anu.edu.au
    }
}

\maketitle
\copyrightnotice
\begin{abstract}
Many NLP researchers rely on free computational services, such as Google Colab, to fine-tune their Transformer models, causing a limitation for hyperparameter optimization (HPO) in long-text classification due to the method having quadratic complexity and needing a bigger resource. In Indonesian, only a few works were found on long-text classification using Transformers. Most only use a small amount of data and do not report any HPO.
In this study, using 18k news articles, we investigate which pretrained models are recommended to use based on the output length of the tokenizer. We then compare some hacks to shorten and enrich the sequences, which are the removals of stopwords, punctuation, low-frequency words, and recurring words. To get a fair comparison, we propose and run an efficient and dynamic HPO procedure that can be done gradually on a limited resource and does not require a long-running optimization library. Using the best hack found, we then compare 512, 256, and 128 tokens length.
We find that removing stopwords while keeping punctuation and low-frequency words is the best hack. Some of our setups manage to outperform taking 512 first tokens using a smaller 128 or 256 first tokens which manage to represent the same information while requiring less computational resources.
The findings could help developers to efficiently pursue optimal performance of the models using limited resources.
\end{abstract}


\section{Introduction}

Classification task datasets may contain long texts. While they potentially have more distinguishable information that can be utilized by trained AI models to more effectively classify them, they are harder and costlier to process than short texts. For a large amount of data, fine-tuning a Transformer~\cite{Vaswani2017Attention} model using long texts is a heavy task that requires a huge computational resource, since the model computes all tokens in parallel and has a quadratic complexity of the self-attention mechanism. This condition discourages many researchers who rely on limited computational services to pursue optimal results using hyperparameter optimization (HPO).

We take a case in the Indonesian research environment as Indonesian is an underrepresented language~\cite{aji2022one, winata-etal-2023-nusax}. The lack of datasets, NLP researchers, and the high computational costs~\cite{nityasya2020costs,nityasya2022student} lead to slow NLP research progress in this area. Many of the researchers who rely on limited computational services, such as Google Colab, face a computational hurdle since the commonly used Transformer-based models~\cite{Koto2020IndoLEM, Wilie2020IndoNLU} are costly, especially when fine-tuned using a large amount of long text data.

In the context of Transformer models, the long text is any text longer than 512 tokens. The early model variants~\cite{Radford2018GPT,Devlin2019BERT} limited the input sequence to 512 tokens due to computational cost. Thus, researchers~\cite{Sun2019BertTextClassification,Pappagari2019Hierarchical,Ding2020CogLTX,Wang2021FeatureSelection,Beltagy2020Longformer,xiong2021nystromformer} tried to find ways to deal with longer text. For the same reason of cost, any text longer than 256 tokens could also be considered as long text, such as news articles.

This observation looks for some simple hacks to lighten up the computational cost; shortening the sequences, speeding up the fine-tuning and the HPO, while improving the performance. The observed alternatives are simple compared to other available solutions~\cite{Pappagari2019Hierarchical,Ding2020CogLTX,Wang2021FeatureSelection} and are assumed as hacks because they are counterintuitive to the contextual pretrained language models and are not recognized as a category of solutions in some surveys~\cite{Park2022Efficient,Qurishi2022Survey}.

Firstly, we investigate the output length of the model's tokenizer to get recommended models. Against 18k news articles of IndoSum~\cite{Kurniawan2018IndoSum}, we find that most tokenizers of Indonesian models in the Huggingface repository~\footnote{https://huggingface.co/models} produce 10\%-14\% more tokens on average, while multilingual models 20\%-48\%. The rest of the Indonesian tokenizers create tokens more than or close to the multilingual ones, which are not recommended to use. Subsequently, we propose a procedure to efficiently tune some of the most important hyperparameters~\cite{Ng2017}, which are the learning rate, the batch size, and the number of epochs. The HPO procedure can be conducted gradually on a time and memory-limited GPU service. It can dynamically find out the optimal F1 score.

Lastly, we investigate simple text-shortening methods which produce better performance. There are several variations to compare: (1) original text, (2) removing stopwords, (3) removing punctuation, (4) removing stopwords and punctuation, (5) removing stopwords and low-frequency words, (6) combining head and tail, and (7) unique words only. The result indicates that removing stopwords outperforms the other methods. We then compare 128-token, 256-token, and 512-token limitations. It turns out that 256-token sequences produce the best result, followed by 128-token. Removing stopwords from the sequences and taking the first 128 or 256 tokens is reliable.

\section{Related Work}\label{section2}
There are limited studies discussing Transformer models over Indonesian long-text classification. All of them use news bodies~\cite{Hutama2022Hoax,Fawaid2021Fake,Juarto2023News,Mahfuzh2022Risk,Mutasodirin2021}. Hutama and Suhartono~\cite{Hutama2022Hoax} study binary fake news detection using 1,100 news articles. Fawaid et al.~\cite{Fawaid2021Fake} study the same, using the same data, with an additional set containing 1,116 news articles. Juarto and Yulianto~\cite{Juarto2023News} study news categorization into 5 classes using 8,754 news articles. All their datasets are publicly available online. Mahfuzh and Purwarianti~\cite{Mahfuzh2022Risk} study multi-label classification with 1,999 news articles and 16 classes, but the dataset is not publicly available online. None of those studies uses k-fold cross-validation for their small datasets, but one~\cite{Mutasodirin2021}.

Our work is inspired by the work of Mutasodirin and Prasojo~\cite{Mutasodirin2021} where IndoSum~\cite{Kurniawan2018IndoSum} is used to compare summarization ground truth against simple truncation in a classification task. So far, it is difficult to find an Indonesian long-text classification dataset larger than IndoSum. Accordingly, we follow them in this study. Since Mutasodirin et al. compare summarization ground truths which are short, our study takes part in longer text. We found no other classification study using IndoSum.

Hutama et al.~\cite{Hutama2022Hoax}, Fawaid et al.~\cite{Fawaid2021Fake}, and Juarto et al.~\cite{Juarto2023News} remove stopwords from the text and keep punctuation. Unfortunately, they neither report the comparison between the text with and without stopwords nor mention other studies to confirm that the method results better. In contrast, Atmajaya et al.~\cite{Atmajaya2022Law} assume oppositely that removing punctuation and keeping stopwords results better. Mahfuzh et al.~\cite{Mahfuzh2022Risk} do not seem to remove stopwords in their experiment. None of them reports a validation of their assumption. Thus, in this paper, we report which proposal gives better performance.

Hutama et al.~\cite{Hutama2022Hoax} and Fawaid et al.~\cite{Fawaid2021Fake} use multilingual models that we avoid when monolingual models are available. Multilingual models use shared vocabulary and the tokenizers produce more tokens~\cite{Conneau2020XLMR,Liang2023XLMV}. Most monolingual models produce lesser tokens so that more information can fit the maximum size of the input. Mahfuzh et al.~\cite{Mahfuzh2022Risk} do not seem to explain what language of BERT they use. Juarto et al. use a certain version of IndoBERT~\cite{Koto2020IndoLEM} and it seems to be the only study we found so far on Indonesian long-text classification using the Indonesian monolingual model.

Most studies mentioned above do not tune hyperparameters, but one~\cite{Hutama2022Hoax}. Comparing some models without tuning hyperparameters seems unfair. Each model needs to meet its best setting to achieve its best performance. However, hyperparameter tuning is so expensive that it needs to be more efficient. Hutama et al.~\cite{Hutama2022Hoax} use Optuna~\cite{Akiba2019Optuna} to look for the optimal hyperparameter setting dynamically. The range of learning rate is 4e-5 until 0.01, weight decay is 4e-5 until 0.01, and epoch is 2 until 5. But still, the search range is manually decided by the researcher in advance which needs prior knowledge of the best range. We provide a different approach for dynamic search in that the best range is decided within the process.

There are many works on making efficient versions of Transformer~\cite{Tay2022Efficient}. The problem is that we found only one Indonesian pretrained version\footnote{https://huggingface.co/ilos-vigil/bigbird-small-indonesian}, but its tokenization output length is close to multilingual models. According to Park, et al.~\cite{Park2022Efficient}, an efficient Transformer that can fit more than 512 tokens, such as Longformer~\cite{Beltagy2020Longformer}, generally still takes a long time to train and certainly has bigger memory relative to 512-token BERT~\cite{Devlin2019BERT} as a standard baseline. Nystromformer~\cite{xiong2021nystromformer} is an exception to this, where they report that their Nystromformer-1024 consumes slightly less memory than standard Transformer-512, but it still takes longer to compute.  Hierarchical methods~\cite{Pappagari2019Hierarchical} also need bigger memory~\cite{Park2022Efficient}. Thus, we avoid suggesting taking tokens of more than 512.

There are also methods to shorten the text, such as summarization techniques or techniques to select key sentences like CogLTX~\cite{Ding2020CogLTX}. However, such techniques take a lot of time to process~\cite{Park2022Efficient} and the result is not better than simple truncation~\cite{Mutasodirin2021} or inconsistent across different datasets~\cite{Park2022Efficient}. Existing simple and fast methods to shorten the text claimed to be the best are taking the document's head only or taking the document's head and tail~\cite{Mutasodirin2021,Sun2019BertTextClassification}. We do not find stopword removal mentioned in long-text classification methods surveys~\cite{Park2022Efficient,Qurishi2022Survey}, although we find many studies in different contexts use this technique.

We close this section by mentioning a Chinese study~\cite{Wang2021FeatureSelection} on classical methods to shorten the text. They use TF-IDF, TextRank, Chi-Square Statistic, and Mutual Information (MI) to score the importance of the words. The shortened texts are then fed into BERT. No HPO is reported in the study. Because the time and the computational resource in our study are limited, and we conduct HPO, we leave the comparison between the best method found in our study and those methods from the Chinese study for future work.

\begin{figure}[t]
\centerline{\includegraphics[width=0.48\textwidth]{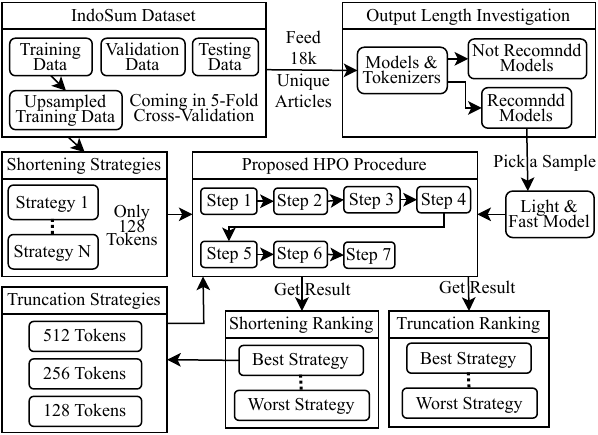}}
\caption{Research Methodology}
\label{fig:methodology}
\end{figure}

\section{Methodology} \label{sec:methodology}
In this section, we explain the dataset, the method to choose good pretrained models to use, the text-shortening strategies to investigate, and the proposed HPO procedure. Figure~\ref{fig:methodology} describes the flow of the methodology.

\subsection{IndoSum as a Classification Dataset}
For the experiment, IndoSum~\cite{Kurniawan2018IndoSum} is used as the benchmark dataset. IndoSum has 18,774 news articles within 6 categories. The biggest class is Headline with 7,192 articles, while the smallest is Inspiration with 130 articles. The rest, Sport, Showbiz, Technology, and Entertainment class, has 4,768, 2,578, 2,303, and 1,803 articles respectively. The dataset comes in 5-fold cross-validation. Each fold contains roughly 3,755 articles and is alternately used as a testing set. The validation set comes from 5\% of the other 4 folds of training data. Thus, the training and validation set consists of roughly 14,268 and 751 articles respectively. For the initial preprocessing, we remove unnecessary whitespaces \textbackslash{n}, \textbackslash{r}, \textbackslash{t}, \textbackslash{xa0}, and space characters.

IndoSum is extremely imbalanced. The models will overfit the Headline class and underfit the others. Bad performance in the Inspiration class drastically decreases F1 score. To cope with this challenge, we avoid new additional data as possible. Instead, we copy the articles of 5 classes in training data until they reach the same amount as the Headline class, around 5.5k articles each, then reshuffle. This way, models can learn from the Inspiration class many times and it prevents overfitting the Headline class too far. The validation and testing sets remain the same. The new 33k-article training data is then referred to as The Upsampled Training Data.

\subsection{Transformer Models and Tokenization Output Length}
There are dozens of Transformer models pretrained using Indonesian corpus. We explore 14 multilingual models and 39 monolingual models from Huggingface\footnote{https://huggingface.co/models}. The models were pretrained by different contributors using the same or different corpora. All those models can be used using a single programming library named Transformers~\cite{Wolf2020Huggingface}.

\textbf{Which models to use?} The first thing to investigate is the tokenization output length. It is to estimate how long the generated input sequence will be. Many Transformer models~\cite{Devlin2019BERT,Koto2020IndoLEM,Lan2020ALBERT,Radford2018GPT,Sanh2019DistilBERT,Wilie2020IndoNLU} limit the input by 512 tokens at most. The more tokens are generated, the more information exceeds the limit. The longer sequences are fed to the network, the heavier the training process will be; due to the quadratic complexity. Therefore, it is important to check out the length of the tokenization output. All 18k unique articles are tokenized using the model's tokenizer.

\subsection{Shortening and Truncation Strategy}
\textbf{Do stopwords and punctuation matter?} The question that arises is whether stopwords and punctuation removal give better performance since some studies do so without comparison reports as discussed in Section~\ref{section2}. We compare some variations: (A.1) normal text, (A.2) text without stopwords, (A.3) text without punctuation, (A.4) text without stopwords and punctuation, (A.5) text without stopwords and low-frequency words, (B.1) head and tail normal text, (B.2) head and tail without stopwords, and (C.1) unique words only. Low-frequency words we remove are the words that appear only once in an article. To make the words unique, we exclude the punctuation and the recurring words.

\textbf{Taking 128, 256, or 512 tokens?} Longer information helps the model. Nevertheless, a question is rising whether the second-time appearing information or the further explanation in the middle and the last part of the article do really matter. Mutasodirin et al.~\cite{Mutasodirin2021} conclude that 75-token sequences give performance close to 512-token sequences. Therefore, we investigate the 128-token and 256-token.

\subsection{Hyperparameter Optimization (HPO) Strategy}
The hyperparameters searching procedure we propose in this study is inspired by Bayesian optimization (BO)~\cite{Bischl2023HPO,Feurer2019HPO,Yu2020HPO}. No parameter is determined in advance. BO is more efficient than grid search and random search. BO should be conducted sequentially because every decision is made based on previous trials. In this proposed method, some processes can be conducted in parallel using several GPU instances if possible. All of the trials are gradually conducted in a time and memory-limited GPU service. This is what the proposed method is intended for, where long-running optimization libraries or software cannot help.

Hyperparameters are classified into two general categories, the ones used to design the model and the ones used to train the model~\cite{Yu2020HPO}. In this study, the architecture design is not modified and left as it is. Among the most important hyperparameters for training are learning rate (LR), batch size (BS), number of epochs, and learning optimizer~\cite{Ng2017}. We tune LR, BS, and the number of epochs. For the optimizer, we only use AdamW~\cite{Loshchilov2018AdamW} and leave its hyperparameters at the default value \(\beta_1\)=0.9, \(\beta_2\)=0.999, \(\epsilon\)=1e-8, and L2 weight decay of 0.01. Due to the GPU quota limitation, we look for the best epoch only in the range of 1-3. Table~\ref{tab-hyper} depicts the setting.

\begin{table}[t]
\centering
\caption{Hyperparameter Setting}
\label{tab-hyper}
\begin{tabular}{ll|ll}
\textbf{Learning Rate:} & \{to search for\} & \textbf{Adam \(\beta_1\):} & 0.9\\
\textbf{Batch Size:} & \{to search for\} & \textbf{Adam \(\beta_2\):} & 0.999\\
\textbf{Epochs:} & [1, 2, 3] & \textbf{Adam \(\epsilon\):} & 1e-8\\
\textbf{Dropout Rate:} & \{default value\} & \textbf{Adam Weight Decay:} & 0.01\\
\textbf{Optimizer:} & AdamW\\
\end{tabular}
\end{table}

Following Mutasodirin et al.~\cite{Mutasodirin2021}, Indonesian DistilBERT-Base\footnote{https://huggingface.co/cahya/distilbert-base-indonesian} is used in this experiment. It is the fastest among the recommended models found. The maximum length of a sequence is limited to 128 tokens. This limitation is useful for some investigations later on. We use Tesla T4 with a memory limit of 16 GB from a free online GPU service. Since this study focuses on HPO, the validation process in every epoch uses testing data to monitor the F1 score and validation data just to monitor the loss. The proposed procedure is explained as follows:

\textbf{Step-1: Reasonable Initial BS, LR, and Number of Epochs.}
The best BS to start with is the biggest size the GPU can take, in the range of mini-batch 16, 32, 64, 128, 256, or 512 samples. The computation runs faster with larger BS and larger LR. Before exploring with the smaller, it is recommended to explore with the bigger to quickly spot the best range. With a maximum memory of 16 GB, the GPU can handle 128 samples per batch for DistilBERT-Base. While for the reasonable LR initial, we start from 5e-5 as suggested by Devlin et al.~\cite{Devlin2019BERT}. Due to the GPU time quota and having more than 255 learning steps in an epoch, we limit the searching range to only 1-3 epochs. BS, LR, and the number of epochs need to be adjusted according to a particular dataset. A smaller dataset requires more epochs or higher LR because it has a smaller batch iteration in an epoch. In this initial step, only one set of IndoSum is utilized. The result of this step is 0.8013, 0.7585, and 0.7794 for epochs 1, 2, and 3 respectively.

\textbf{Step-2: Understanding the Flow Direction.}
The objective of this optimization hack is to efficiently assess the LRs resulting in (1) a smooth transition of F1 score between epochs, meaning that each epoch gets optimal learning steps, and (2) giving the highest average of all sets. From the result of Step-1, the direction should flow to the lower LRs because LR for the 1st epoch is too high, resulting in a great degradation in the 2nd epoch. The next LRs to investigate could be 1e-5, 5e-6, and 1e-6, using only one set of IndoSum. Table~\ref{tab-step1-2} depicts the result.

\begin{table}[t]
\centering
\caption{F1 Scores in Step-1 and Step-2. DistilBERT uses only the 1st set of IndoSum with a BS of 128. The result indicates that the best LRs are around 5e-6 or 1e-5.}
\label{tab-step1-2}
\begin{tabular}{r|cccc}
 & \multicolumn{4}{c}{\textbf{LR}}\\
\textbf{Epoch} & \textbf{1e-6} & \textbf{5e-6} & \textbf{1e-5} & \textbf{5e-5}\\
\cline{2-5}
\textbf{1st} & 75.78 & 80.88 & 80.66 & 80.13\\
\textbf{2nd} & 78.72 & 81.83 & 81.92 & 75.85\\
\textbf{3rd} & 79.96 & 82.03 & 79.79 & 77.94\\
\end{tabular}
\end{table}

\textbf{Step-3: Deciding the Best Range.} The result of Step-2 tells that the best LRs lie around LR=5e-6 or LR=1e-5. The LR=1e-6 indicates that 3-epoch training is not sufficient and the F1 score does not peak in the 3rd epoch. In this step, the average F1 score from all sets of IndoSum is calculated for LR 5e-6 and 1e-5. The results are 0.8216, 0.8113, 0.8014 for LR=1e-5 and 0.8078, 0.8192, 0.8205 for LR=5e-6. The average scores tell more about the range. The best LR for the 3rd epoch lies around LR=5e-6 and the best LR for the 1st epoch lies around LR=1e-5.

\textbf{Step-4: First Propagation.} The first propagation, using all sets of IndoSum, starts from LR=5e-6 to the lower and to the higher LR. The objective is to look for the best average F1 score for the 3rd epoch. The results are 0.8171 for LR=4e-6 and 0.8046 for LR=6e-6. Since the result of LR=4e-6 at the 3rd epoch is worse than LR=5e-6, the propagation stops. The same with LR=6e-6. The conclusion of the first propagation is that LR=5e-6 is the best for the 3rd epoch.

\textbf{Step-5: Second Propagation.} The objective of this second propagation is to find the best LR for the 2nd epoch. From the result of the first propagation, see Table~\ref{tab-step3-7}, the second propagation starts from LR=4e-6 to another neighbor because it has the highest F1 score which is 0.8194. Using all sets of IndoSum, the trials are done in only 2 epochs. The results are 0.8152 for LR=3e-6. The conclusion of the second propagation is that LR=4e-6 is the best for the 2nd epoch.

\textbf{Step-6: Third Propagation.} The objective of the third propagation is to find the best LR for the 1st epoch. From the result of the previous steps, the third propagation starts from LR=1e-5 because it has the highest F1 score which is 0.8216. Using all sets of IndoSum, the trials are done in only 1 epoch. The results are 0.8179 for LR=9e-6 and 0.8063 for LR=2e-5. The conclusion of the third propagation is that LR=1e-5 is the best for the 1st epoch.

\textbf{Step-7: Trying on Smaller BS.} Smaller BS requires the same or lower LR because it has more learning steps to complete an epoch. The propagation flows to the lower LRs. Table~\ref{tab-step3-7} depicts the result. Finally, how deep the optimization is depends on the user's needs and limitations. Users can also try another learning optimizer using the best hyperparameter range found before.

\begin{table*}[t]
\centering
\caption{Average F1 Scores from Step-3 to Step-7. DistilBERT uses all sets of IndoSum.}
\label{tab-step3-7}
\begin{tabular}{lr|ccccccccccccc}

 & & & & & & & & \textbf{LR}\\

 & \textbf{Epoch} & \textbf{8e-7} & \textbf{9e-7} & \textbf{1e-6} & \textbf{2e-6} & \textbf{3e-6} & \textbf{4e-6} & \textbf{5e-6} & \textbf{6e-6} & \textbf{7e-6} & \textbf{8e-6} & \textbf{9e-6} & \textbf{1e-5} & \textbf{2e-5}\\
\cline{3-15}

 & \textbf{1st} & - & - & - & - & 79.19 & 80.04 & 80.78 & 81.41 & - & - & 81.79 & \textbf{82.16} & 80.63\\
\textbf{BS=128} & \textbf{2nd} & - & - & - & - & 81.52 & \textbf{81.94} & 81.92 & 81.01 & - & - & - & 81.13 & -\\
 & \textbf{3rd} & - & - & - & - & - & 81.71 & \textbf{82.05} & 80.46 & - & - & - & 80.14 & -\\
\\

 & & \textbf{8e-7} & \textbf{9e-7} & \textbf{1e-6} & \textbf{2e-6} & \textbf{3e-6} & \textbf{4e-6} & \textbf{5e-6} & \textbf{6e-6} & \textbf{7e-6} & \textbf{8e-6} & \textbf{9e-6} & \textbf{1e-5} & \textbf{2e-5}\\
\cline{3-15}

 & \textbf{1st} & - & - & 77.37 & 79.03 & 80.33 & 81.37 & - & 80.97 & \textbf{82.18} & 82.01 & 81.39 & - & - \\
\textbf{BS=64} & \textbf{2nd} & - & - & 79.55 & \textbf{81.73} & 81.32 & 81.55 & - & - & - & - & - & - & -\\
 & \textbf{3rd} & - & - & 80.61 & \textbf{81.67} & 81.47 & 80.69 & - & - & - & - & - & - & -\\
\\

 & & \textbf{8e-7} & \textbf{9e-7} & \textbf{1e-6} & \textbf{2e-6} & \textbf{3e-6} & \textbf{4e-6} & \textbf{5e-6} & \textbf{6e-6} & \textbf{7e-6}\\
\cline{3-15}

 & \textbf{1st} & 77.84 & 78.05 & 78.27 & 80.37 & - & - & - & 81.50 & \textbf{81.53} & - & - & - & -\\
\textbf{BS=32} & \textbf{2nd} & 79.74 & 80.28 & 80.80 & \textbf{81.87} & - & - & - & - & - & - & - & - & -\\
 & \textbf{3rd} & 81.03 & \textbf{81.62} & 81.47 & 81.36 & - & - & - & - & - & - & - & - & -\\
\end{tabular}
\end{table*}

\section{Result and Discussion} \label{sec:discussion}
Table~\ref{tab-tokenization} depicts the investigation result of tokenization output length. On average, the multilingual tokenizers produce 20\% or more tokens. The most-vocabulary multilingual tokenizer results in 7\% additional length at best. Monolingual tokenizers should have lesser. Around 26 of 39 Indonesian Transformer models investigated produce 10\%-14\% more tokens on average and may produce 0\% at best, meaning that there are articles of which the models know all of the words and none of them is broken down into subwords. Upon this result, we recommend monolingual models having a minimum of no more than 0\% and an average of no more than 15\%, making it better than multilingual models.

We cannot say that the 10\% is always better than the 15\%. While a model has more vocabulary and produces lesser additional tokens, more vocabulary needs bigger training data to teach the model and a bigger network to represent the language understanding~\cite{Conneau2020XLMR,Liang2023XLMV}. The bigger model could be a problem in a low-memory GPU. There will be a trade-off between the vocabulary size and the model size. Thus, our recommendation is in range, not one specific model.

\begin{table}[t]
\centering
\caption{Result Summary on Additional Length (in \%) of Sequence after Tokenization Process}
\label{tab-tokenization}
\begin{tabular}{lccc}
\hline
\multicolumn{4}{c}{Of 26 Recommended Monolingual Models}\\
\hline
& \multicolumn{3}{c}{\textbf{Additional Length}} \\
\textbf{Description} & \textbf{Max.} & \textbf{Min.} & \textbf{Avg.}\\
The Most-Vocabulary Model & \textbf{70\%} & \textbf{0\%} & \textbf{10\%}\\
The Least-Vocabulary Model & 85\% & 0\% & 14\%\\
\\
\hline
\multicolumn{4}{c}{Of 14 Multilingual Models}\\
\hline
& \multicolumn{3}{c}{\textbf{Additional Length}} \\
\textbf{Description} & \textbf{Max.} & \textbf{Min.} & \textbf{Avg.}\\
The 1st Most-Vocabulary Model & \textbf{79\%} & \textbf{7\%} & \textbf{20\%}\\
The 2nd Most-Vocabulary Model & 112\% & 9\% & 35\%\\
The Least-Vocabulary Model & 131\% & 16\% & 48\%\\
\end{tabular}
\end{table}

Table~\ref{tab-short} depicts the result of the shortening method comparison. It shows that removing stopwords gives the best improvement. It is because more information fits 128 tokens limitation and the model still understands the text. Punctuation helps the model to understand the context and removing them degrades the performance. Classifying sequences of unique words works well because the initial sentences are understandable. While stopwords do not provide useful information, low-frequency words do not represent the topic. However, removing low-frequency words is harmful. Combining the head and the tail of the article does not give a better result.

\begin{table}[t]
\centering
\caption{Shortening Strategy Comparison. An experiment using DistilBERT with a maximum of 128 Tokens.}
\label{tab-short}
\begin{tabular}{lcccc}
\textbf{Strategy} & \textbf{BS} & \textbf{LR} & \textbf{Epochs} & \textbf{F1}\\
\hline
A.2 & 16 & 9e-7 & 3 & \textbf{83.01}\\
C.1 & 16 & 3e-6 & 2 & 82.61\\
A.4 & 64 & 2e-6 & 3 & 82.49\\
A.1 (baseline 1) & 16 & 4e-6 & 1 & 82.31\\
A.3 & 128 & 5e-6 & 2 & 82.29\\
B.2 & 16 & 1e-6 & 3 & 82.18\\
B.1 (baseline 2) & 64 & 8e-6 & 1 & 81.54\\
A.5 & 64 & 7e-6 & 1 & 80.41\\
\end{tabular}
\end{table}

Table~\ref{tab-short2} depicts the result of the truncation method comparison. We find that 256-token and 128-token sequences give better performance than 512-token sequences. Although the model can only take at most 32 samples per batch for 512-token sequences, by a fair comparison using BS=32 and BS=16, it turns out that giving too long text might decrease the performance. For limited computing resources, avoiding the full 512 tokens and taking 128 or 256 tokens is reliable.

\begin{table}[t]
\centering
\caption{Truncation Strategy Comparison. An experiment using DistilBERT with strategy A.2.}
\label{tab-short2}
\begin{tabular}{ccccc}
\textbf{Tokens} & \textbf{BS} & \textbf{LR} & \textbf{Epochs} & \textbf{F1}\\
\hline
256 & 16 & 1e-6 & 3 & \textbf{83.02}\\
128 & 16 & 9e-7 & 3 & \textbf{83.01}\\
512 & 16 & 2e-6 & 2 & 82.93\\
\hline
256 & 32 & 6e-6 & 1 & \textbf{82.81}\\
256 & 64 & 9e-6 & 1 & \textbf{82.77}\\
128 & 64 & 3e-6 & 2 & \textbf{82.70}\\
512 & 32 & 9e-7 & 3 & 82.47\\
\end{tabular}
\end{table}

According to Table~\ref{tab-step3-7}, \ref{tab-short}, and \ref{tab-short2}, training in only one epoch could be better. Although some studies~\cite{Devlin2019BERT,Hutama2022Hoax} do not recommend one-epoch training, it is not a consensus. It all depends on the dataset and the model. So do the batch size and the learning rate, no consensus for the best value.

\section{Conclusion and Future Work} \label{sec:conclusion}

We investigate and propose some strategies to (1) lighten up the fine-tuning while improving the performance and (2) speed up the HPO in a time and memory-limited GPU service. We recommend using Indonesian model whose tokenizer produces less than 15\% additional tokens on average. Our HPO procedure consists of 7 steps and can be conducted in a gradual manner. Shortening and enriching the sequences by removing stopwords while keeping punctuation and low-frequency words works well to improve performance. It also turns out that too long the text does not always give better results and might decrease performance.

There are some open problems emerging. The first is about the trade-off between vocabulary size and model size. The second is the need for efficient models in Indonesian. The third is a better text-shortening strategy. The fourth is finding better models on the IndoSum classification dataset. It seems that IndoSum classification is still challenging, due to the data imbalance. We leave all of these problems to future work.

\bibliographystyle{IEEEtran}
\bibliography{IEEE_conference}

\end{document}